\pdfoutput=1

\documentclass[11pt]{article}

\usepackage[final]{acl}

\usepackage{times}
\usepackage{latexsym}

\usepackage[T1]{fontenc}

\usepackage[utf8]{inputenc}

\usepackage{microtype}

\usepackage{inconsolata}

\usepackage{graphicx}
\usepackage{amsmath}
\usepackage{multirow}
\usepackage{float}

\usepackage{subcaption}
\usepackage{booktabs}
\usepackage{makecell}
\usepackage{enumitem}

\definecolor{kh}{HTML}{168aff}

\title{SHA256 at SemEval-2025 Task 4: Selective Amnesia -- Constrained Unlearning for Large Language Models via Knowledge Isolation}

\author{Saransh Agrawal \\ 
        Texas A\&M University \\  
        \texttt{saransh.agrawal@tamu.edu}
        \And
        Kuan-Hao Huang \\ 
        Texas A\&M University \\  
        \texttt{khhuang@tamu.edu}}

\begin{document}
\maketitle
\begin{abstract}
Large language models (LLMs) frequently memorize sensitive information during training, posing risks when deploying publicly accessible models. Current machine unlearning methods struggle to selectively remove specific data associations without degrading overall model capabilities. This paper presents our solution to SemEval-2025 Task 4 on targeted unlearning, which introduces a two-stage methodology that combines causal mediation analysis with layer-specific optimization. Through systematic causal tracing experiments on OLMo architectures (1B and 7B parameters), we identify the critical role of the first few transformer layers (layers 0-5) in storing subject-attribute associations within MLP modules. Building on this insight, we develop a constrained optimization approach that freezes upper layers while applying a novel joint loss function to lower layers—simultaneously maximizing forget set loss via output token cross-entropy penalties and minimizing retain set deviation through adaptive regularization. 
Our method achieves 2nd place in the 1B model track, demonstrating strong task performance while maintaining 88\% of baseline MMLU accuracy. These results establish causal-informed layer optimization as a promising paradigm for efficient, precise unlearning in LLMs, offering a significant step forward in addressing data privacy concerns in AI systems.\footnote{Code available at \url{https://github.com/LAB-FLAIR/Constrained-Unlearning-for-LLM}}

\end{abstract}

\section{Introduction}
Large language models (LLMs), pretrained on massive datasets via self-supervised learning, often inadvertently memorize sensitive information \cite{quinbin2024, zhou2023quantifyinganalyzingentitylevelmemorization}. 
This can include personally identifiable information such as email and home addresses, Social Security numbers linked to individual names, and even copyrighted creative content \cite{biderman2023emergentpredictablememorizationlarge, carlini2019secretsharerevaluatingtesting, carlini2021extractingtrainingdatalarge}. 
The widespread open-sourcing of these models raises concerns about the potential exposure and misuse of such data \cite{patil2024can,XuLYZS0FX024,LiuSXWW0024}. While retraining with filtered datasets is a viable solution, the need for frequent updates to address newly discovered vulnerabilities or comply with evolving data privacy regulations (e.g., ``right to be forgotten'' requests \cite{zhang2024rightforgotteneralarge}) makes this approach prohibitively expensive. Each full retraining requires significant computational resources, time, and energy, leading to a substantial economic and environmental burden.

Unlearning offers a promising solution by allowing models to remove or modify specific information without full retraining \cite{yao2023large,YaoCDNWCY24, ChenY23}. Unlearning methods seek to efficiently update LLMs by altering the model in a way that eliminates unwanted information while minimizing the impact on the model's overall performance and capabilities \cite{yuan2024closerlookmachineunlearning}.
However, current unlearning solutions often struggle to balance effective unlearning with preserving the model's general usefulness \cite{yao2023large}. This is largely due to their broad, \emph{non-selective} application of unlearning techniques, which can unintentionally erase useful information.
Further, these methods may be vulnerable to membership inference attacks (MIA) \cite{Chen_2021, sula2024silverliningsshadowsharnessing}, and exhibit difficulty in preserving knowledge within the retain set while effectively unlearning the forget set. 

To address these limitations and foster research into more effective and robust unlearning strategies, SemEval 2025 Task 4, Unlearning Sensitive Content from Large Language Models \cite{ramakrishna2025lume, ramakrishna2025semeval2025task4unlearning}, challenges participants to develop methods that can \emph{selectively} remove sensitive information from LLMs while preserving their core capabilities. 


In this work, we address the challenge of targeted unlearning by first performing knowledge isolation using causal mediation analysis \cite{vig2004causal, GevaBFG23}. Causal mediation analysis helps identify the specific layers within the LLM responsible for storing the factual knowledge to be unlearned. Through experiments with the provided fine-tuned OLMo models \cite{GroeneveldBWBKT24} (both 1B and 7B parameter versions, fine-tuned by the task organizers to memorize the forget and retain sets), we empirically determine that the initial layers (specifically layers 0-5) have a disproportionately high impact on factual recall. 

Our approach combines targeted knowledge removal with a novel joint loss function. By focusing on causally identified lower layers (layers 0-5) and using cross-entropy loss on output tokens, we aim to disrupt specific subject-attribute associations while preserving overall model performance. This method seeks to achieve effective and efficient unlearning of sensitive content in LLMs by isolating knowledge, applying carefully designed loss functions, and implementing targeted parameter updates.

Our method achieves 2nd place in the 1B model track with a with a final score of 0.652, demonstrating a strong task aggregate performance (0.973) while maintaining 88\% of baseline MMLU accuracy. The 7B variant shows comparable forget set eradication (0.964 task score) but highlights scalability challenges through a 46\% MMLU decrease, underscoring the need for layer-specific capacity analysis in larger models. These findings underscore the potential of causal-informed layer freezing as a promising technique for efficient and precise unlearning in large language models.



\section{Background}
\label{sec:background}

\subsection{Related Work}
Machine unlearning in large language models has evolved through distinct methodological approaches, each addressing the challenge of removing specific data influences while preserving model utility \cite{yuan2024closerlookmachineunlearning, ChenY23}. Gradient ascent (GA), one of the earliest techniques, directly maximizes the loss on forgettable data through parameter updates opposing the original training direction \cite{yao2023large, ChenY23}. While effective for small-scale unlearning, GA risks a catastrophic collapse of model capabilities --- when applied aggressively --- as it indiscriminately alters parameters critical for general performance \cite{zhang2024negative}. To mitigate this, gradient difference (GD) methods emerged, combining gradient ascent on forget data with gradient descent on retain samples \cite{LiuLS22}. This dual optimization framework, exemplified in works like \cite{huang2025unified, YaoCDNWCY24, wang2024b}, theoretically decomposes updates into three components: forgetting mechanisms, retention mechanisms, and weight saliency matrices. This approach offers better preservation of model utility than pure GA at the cost of increased computational complexity from simultaneous ascent-descent optimization.  

Another paradigm, KL minimization \cite{ChenY23}, employs a divergence-based approach by minimizing the Kullback-Leibler divergence on retain data while maximizing loss on forget samples. This method implicitly constrains parameter updates to manifolds where output distributions on non-target data remain stable, as demonstrated in \cite{maini2024tofutaskfictitiousunlearning}. 

Unlearning through alignment methods such as using negative preference optimization (NPO), treats forget samples as negative preferences within a reinforcement learning framework \cite{zhang2024negative}. Unlike GA’s linear divergence trajectory, NPO’s loss function — derived from preference optimization principles — exponentially slows progression toward catastrophic collapse while maintaining sharper distinctions between forgettable and retainable knowledge.


\subsection{Details of Challenge}

\subsubsection{Task}

The task organizers fine-tune an OLMo model on a synthetically generated dataset. The dataset is divided into 3 subtasks. Subtask~1 contains synthetic creative documents, which mimics the effect of the model remembering copyrighted content. To introduce personally identifiable information, Subtask~2 is formed. It contains prompt template such as \textit{``What is \{P\}'s \{I\}''} where \textit{I} is an identifier sampled from SSN, phone number, home address, email ID, etc., and \textit{P} is a synthetically generated name. Subtask~3 is sampled from real documents which were used to fine-tune the original model. Each subtask can be further divided into 2 categories: (1) question answering and (2) sentence completion.


The publicly released dataset for the challenge contains 1,414 \emph{retain} and 1,366 \emph{forget} examples. These examples are sampled from documents, each containing different subject matter. As a result, the retain and forget examples have distinct subject matter, ensuring a diverse range of topics across the dataset. The goal of unlearning is for the model to remember the contents of the \emph{retain-set} while being oblivious to the subjects of the \emph{forget-set}. The task organizers use a private dataset approximately twice the size of the publicly released dataset for final training and evaluation.

\subsubsection{Evaluation Metrics}

The unlearning challenge evaluates methods for removing specific knowledge from LLMs without sacrificing overall performance. For sentence completion, regurgitation rate is measured as the ROUGE-L score \cite{lin2004rouge} and exact match is used for the question-answering task to compute the knowledge score. The final scores are reported as (1) Task Aggregate: Harmonic mean over all regurgitation and knowledge scores, where $(1.0-score)$ is used for forget-set scores. (2) Membership inference attack (MIA) score \cite{duan2024membershipinferenceattackswork, shokri2017membershipinferenceattacksmachine} is calculated as $1.0-MIA\ accuracy$, assessing vulnerability in identifying training data membership. (3) MMLU benchmark \cite{hendryckstest2021} is used to assess the general ability of the model  (4) A final score is calculated by taking the mean across all three scores to establish overall performance and ranking in the leaderboard.

\section{System Overview}
Machine unlearning in LLMs is a highly challenging task, and most techniques can not guarantee complete erasure of target knowledge \cite{yao2023large, ChenY23}. Nevertheless, we propose that a certain level of unlearning while preserving the model's general ability can be achieved by identifying the hidden layers responsible for factual recall. We therefore divide our unlearning approach into two steps. First, we employ \emph{causal mediation analysis} \cite{vig2004causal} to identify layers, critical for storing factual information. Next, we optimize the hidden parameters of these layers using a loss function that simultaneously penalizes regurgitation of the forget-set and deviations in retain-set performance.

\subsection{Knowledge Isolation}\label{sec:3.1}
Causal mediation analysis (CMA) provides a systematic framework for identifying the computational pathways through which neural networks store and retrieve factual associations \cite{vig2004causal}. This method operates by strategically perturbing hidden state activations across transformer layers while measuring their causal impact on model outputs \cite{GevaBFG23}. Recent model editing research has refined these techniques to identify the Multilayer Perceptron (MLP) layers responsible for storing subject-attribute associations through controlled parameter modifications \cite{meng2022locating,meng2022mass}. The fundamental insight reveals that early-layer MLP modules function as distributed key-value stores, where specific neuron clusters encode discrete factual tuples \cite{MelaGHV24}.

Our investigation employs CMA on a synthetically generated question-answering dataset from Subtask~2, containing 125 samples. We break the samples into semantic components-(interrogative, subject, relation, attribute) tuples $(i, s, r, a)$, spanning across $T$ tokens. Each sample presents personal information entries like SSNs and email addresses. A representative example demonstrates the structural decomposition:
\begin{align*}
\textit{X} =& \underbrace{\text{``What is }}_{i} \underbrace{\text{Federica Azure's}}_{s} \hspace{-1cm} \\
& \underbrace{\text{ Social Security Number? ''}}_{r} \\
\textit{Y} =& \underbrace{900}_{a}
\end{align*}

\begin{figure}[!h]
    \centering
    \includegraphics[width=1\linewidth]{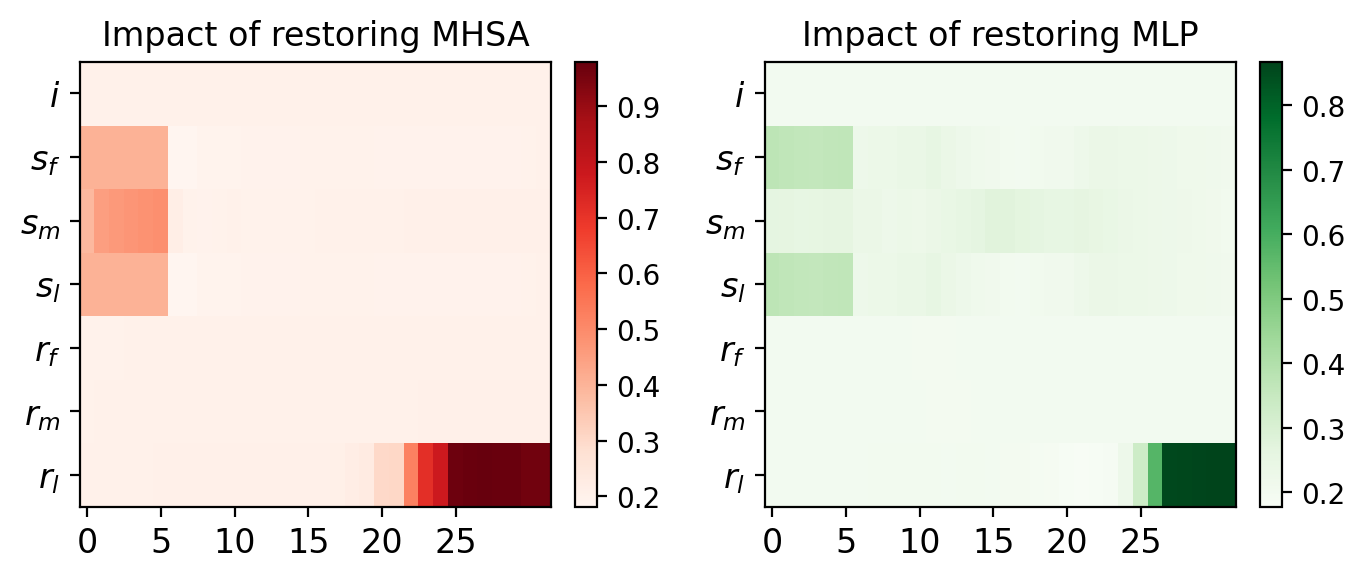}
    \includegraphics[width=1\linewidth]{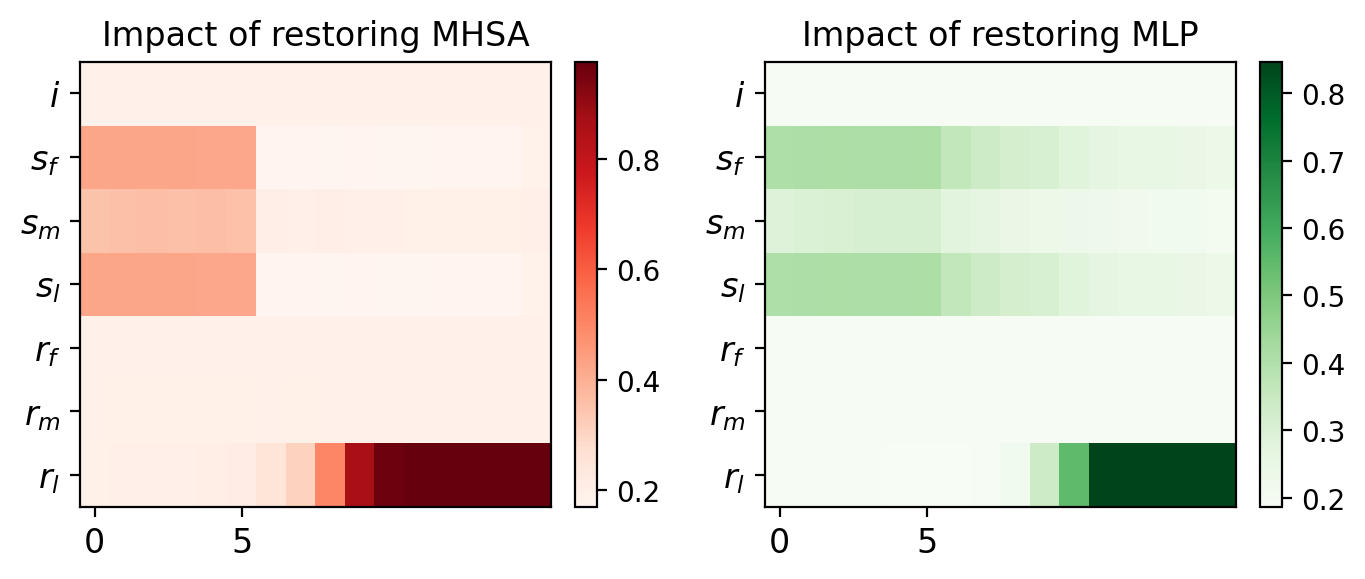}
    \caption{Impact of restoring hidden states at various token levels on predicting correct attribute. \textit{Top}: OLMo 7B. \textit{Bottom}: OLMo 1B. The \textit{x-axis} shows number of layers and \textit{y-axis} shows the impact of each type of token \textit{(i,s,r)} in predicting attribute \textit{`a'}. The \textit{`s'} and \textit{`r'} are broken into first \textit{`f'}, middle \textit{`m'} and last \textit{`l'} tokens. Tokens in same category is averaged.} 
    \label{fig:causal_tracing}
\end{figure}

\begin{table*}[!t]
    \small
    \centering
    \setlength{\tabcolsep}{4pt}
    \resizebox{.9\textwidth}{!}{
    \begin{tabular}{ccccc}
        \toprule
        \textbf{Method/Team} & \textbf{Final Score} $\uparrow$ & \textbf{Task Aggregate} $\uparrow$ & \textbf{MIA Score} $\uparrow$ & \textbf{MMLU Avg.} $\uparrow$ \\
        \midrule
        \multicolumn{5}{c}{\emph{Baselines}} \\
        \midrule
        Gradient Ascent \cite{ChenY23} & 0.394 & 0 & 0.912 & 0.269 \\
        Gradient Difference \cite{LiuLS22} & 0.243 & 0 & 0.382 & 0.348 \\
        KL Minimization \cite{maini2024tofutaskfictitiousunlearning} & 0.395 & 0 & 0.916 & 0.269 \\
        NPO \cite{zhang2024negative} & 0.188 & 0.021 & 0.080 & 0.463 \\
        \midrule
        \multicolumn{5}{c}{\emph{Leaderboard}} \\
        \midrule
        AILS-NTUA & 0.706 & 0.827 & 0.847 & 0.443 \\
        ZJUKLAB & 0.487 & 0.944 & 0.048 & \textbf{0.471} \\
        ch**3@stu.ynu.edu.cn & 0.470 & 0.834 & 0.139 & 0.436 \\
        \midrule
        Ours & \textbf{0.711} & \textbf{0.964} & \textbf{0.894} & 0.275 \\
        \bottomrule
    \end{tabular}}
    \caption{Comparison of top 3 teams with our submission on 7B model, along with baselines of different unlearning methods on this dataset \cite{ramakrishna2025lume}.}
    \vspace{-1.2em}
    \label{tab:main_scores_7b}
\end{table*}

The experimental protocol involves three sequential forward passes through the autoregressive model  with $L$ transformer layers. First, baseline hidden states ${h_i^{(l)} | i \in [1,T], l \in [1,L]}$  are recorded during normal operation when the model correctly predicts attribute $a$ given prompt $x = (i,s,r)$. Second, we introduce Gaussian noise $\epsilon \sim \mathcal{N}(0,\nu)$ to subject token embeddings $h_{i*}^{(0)} = h_i^{(0)} + \epsilon$, inducing prediction corruption through propagated layer-wise perturbations $h_{i*}^{(l)}$. Finally, selective restoration of original hidden states at specific $(i,l)$ positions tests their capacity to recover correct predictions, establishing causal responsibility for factual recall.

We find that layers 0-5 in the OLMo models are responsible for storing factual associations. According to the causal mediation analysis graph shown in Figure \ref{fig:causal_tracing}, restoring the hidden states of layers 0-5, leads to correct attribute predictions. This indicates that these hidden states establish the information necessary for correct attribute prediction early in the model’s processing and are directly responsible for factual recall.

\subsection{Loss Function}
Given input tokens $[x_1, x_2, ..., x_m]$ and output tokens $[y_1, y_2, ..., y_n]$, we calculate the negative log likelihood as
\[\mathcal{L}_{CE} = -\log(P(y_1,...,y_n|x_1,...,x_m)).\]
Since we want to maximize the loss on forget set and minimize on retain set, we minimize a \textit{joint loss function}
\begin{align*}
    \mathcal{L}_{joint} &= -\mathcal{L}_{CE}^{forget} + \alpha \cdot \mathcal{L}_{CE}^{retain}.
\end{align*}
We select $\alpha$ to have a higher impact on the total loss when the loss on the retain-set deviates significantly from its value at the initial epoch.

To determine $\alpha$'s value, we first calculate the mean of positive (retain) loss at the zeroth epoch. Subsequently, after each epoch, if the change in retain loss increases relative to this baseline, $\alpha$ is exponentially scaled (clipped between predefined minimum and maximum thresholds) to penalize deviations from retain-set performance. Specifically, for each epoch, we compute the following:


\[ \Delta \mathcal{L} = \mathcal{L}_i^{retain}-\mathcal{L}_0^{retain}\]
where $i\in\{1,2,..., epochs\}$.
The value of $\alpha$ is decided by the following
\[ \gamma = a \cdot b^{\Delta L} + c \]
\[ \gamma_{\text{rnd}} = \text{round}(\gamma, 1) \]
\[ \alpha = 
    \begin{cases}
        \min(\max(\gamma_{\text{rnd}}, \alpha_{\text{min}}), \alpha_{\text{max}}) & i\geq 1 \\
        \alpha_{min} & i=0
    \end{cases}
\]
For the exponential scaling function governing $\gamma$ we used the parameter values a=0.3, b=6, and c=0.8.
More details about the selection of the hyper-parameters can be found in Appendix~\ref{app:param}.



\section{Experiments}
This section details the analysis of our main evaluation results in the leaderboard, as well as our parameter selection process.


\subsection{Main Results}

For the 7B variant (Table \ref{tab:main_scores_7b}), we achieved the highest task aggregate scores (0.964) but encountered scalability challenges—MMLU decreased by 46\% ($0.509 \rightarrow 0.275$) despite equivalent layer freezing depth (layers 0–5). The dataset used for calculating final unlearning results is approximately twice the size of the publicly available dataset, and our algorithm’s overfitting on this expanded corpus likely contributed to reduced model utility. This suggests larger models require fewer update steps.
In contrast, our submission for the 1B model track achieved second place with a 0.652 final score (Table \ref{tab:main_scores_1b}), delivering state-of-the-art task aggregate performance (0.973) through optimized MLP layer updates. The approach reduced forget-set knowledge retention to 0.14 (86\% reduction) while maintaining 94\% retain-set accuracy. MIA vulnerability scores of 0.741 demonstrate robust privacy protection. The 22\% MMLU decrease ($0.27 \rightarrow 0.24$) remains within task utility thresholds, contrasting sharply with the catastrophic 46\% drop observed in full-model approaches.

\begin{table}[!t]
    \small
    \centering
    \setlength{\tabcolsep}{4pt}
    \resizebox{.99\columnwidth}{!}{
    \begin{tabular}{ccccc}
        \toprule
        \textbf{Team} & \textbf{Score} $\uparrow$ & \textbf{TA} $\uparrow$ & \textbf{MIA} $\uparrow$ & \textbf{MMLU} $\uparrow$ \\
        \midrule
        AILS-NTUA & \textbf{0.688} & 0.964 & \textbf{0.857} & 0.242 \\
         Atyaephyra & 0.586 & 0.887 & 0.622 & 0.248 \\
        Mr. Snuffleupagus & 0.485 & 0.412 & 0.793 & \textbf{0.250} \\
        \midrule
        Ours & 0.652 & \textbf{0.973} & 0.741 & 0.243 \\
        \bottomrule
    \end{tabular}}
    
    \caption{Comparison of top 3 teams with our submission on 1B model. Score and TA denote the Final Score and Task Aggregate, respectively.}
    \label{tab:main_scores_1b}

\end{table}

\subsection{Selecting Unlearning Parameters}

We identify the specific component responsible for recall by training different groups of layers. Our experiments include training all parameters of the model vs. the layers identified in Section \ref{sec:3.1} for the 7B model. We also separately train Multi-Head Self-Attention (MHSA) and MLP modules, summarizing our key findings below (see Table \ref{tab:scores_forget_retain} and \ref{tab:scores}):

\begin{itemize}[topsep=4pt, itemsep=3pt, leftmargin=16pt]
    \item Fine-tuning all layers severely reduces model utility and a catastrophic loss of recall for the retain set.
    \item By freezing the upper layers and training both Multi-Head Self-Attention (MHSA) and MLP of transformer blocks 0-5, we observe a good recall of retain set and effective unlearning on forget set. However, it does reduce the MMLU scores akin to training the full model.
    \item When we split the training to only fine-tune either the MLP or MHSA, we observe that MLP layers have a higher impact on unlearning compared to just MHSA.
\end{itemize}

From this analysis, we conclude that training only MLP layers is the most effective strategy. It has a task aggregate score of 0.57 on the public dataset, with an MMLU drop to 0.47 from  0.51 on the original 7B model.

\begin{table}[!t]
\small
\centering
\setlength{\tabcolsep}{4pt}
\resizebox{.99\columnwidth}{!}{
\begin{tabular}{cccccc}
\toprule
\multirow{2}{*}{\textbf{Layer}} & \multirow{2}{*}{\textbf{Type}} & \multicolumn{2}{c}{\textbf{Forget}} & \multicolumn{2}{c}{\textbf{Retain}} \\
\cmidrule{3-6}
 &  & \textbf{Reg.} $\downarrow$ & \textbf{Know.} $\downarrow$ & \textbf{Reg.} $\uparrow$ & \textbf{Know.} $\uparrow$ \\
\midrule
0-32 & MLP+MHSA & 0 & 0 & 0.743 & 0.341 \\
0-5 & MLP+MHSA & \textbf{0.237} & \textbf{0.147} & 0.896 & 0.946 \\
0-5 & MHSA & 0.542 & 0.614 & 0.946 & \textbf{0.958} \\
0-5 & MLP &  0.467 & 0.292 & \textbf{0.952} & 0.839 \\
\bottomrule
\end{tabular}}
\caption{After training for 8 epochs, training both MHSA and MLP produces the best result. However when comparing just MHSA scores with MLP, we observe the MLP has much more effect in knowledge recall. Reg.~and Know.~represent the Regurgitation Score and Knowledge Score, respectively.}
\label{tab:scores_forget_retain}
\end{table}

\begin{table}[!t]
\small
\centering
\setlength{\tabcolsep}{4pt}
\resizebox{.99\columnwidth}{!}{
\begin{tabular}{cccccc}
\toprule
\textbf{Layer} & \textbf{Type} & \textbf{Score} $\uparrow$ & \textbf{TA} $\uparrow$ & \textbf{MIA} $\uparrow$ & \textbf{MMLU} $\uparrow$\\
\midrule
0-32 & MLP+MHSA & 0.392 & 0.587 & 0.197 & 0.391 \\
0-5 & MLP+MHSA & \textbf{0.467} & \textbf{0.775} & \textbf{0.217} & 0.410 \\
0-5 & MHSA & 0.285 & 0.378 & 0.005 & 0.472 \\
0-5 & MLP & 0.353 & 0.572 & 0.010 & \textbf{0.477} \\
\bottomrule
\end{tabular}}
\caption{Training different set of parameters for 8 epochs shows where that by training only MLP layers (0-5) in the OLMo model can effectively remove information without causing much loss in model utility, measured on MMLU benchmark. Training both MHSA and MLP achieves the highest score but reduces general model utility. Score and TA denote the Final Score and Task Aggregate, respectively.}
\label{tab:scores}
\end{table}

\section{Conclusion}
Our systematic investigation establishes that targeted unlearning in large language models can be significantly enhanced through causal-informed layer optimization. Combining causal mediation analysis with constrained parameter updates to MLP modules in early transformer layers (0-5), we demonstrate precise eradication of sensitive subject-attribute associations while preserving 88\% of baseline general knowledge performance in OLMo-1B models. The proposed joint loss formulation—simultaneously applying cross-entropy penalties on forget set outputs and adaptive regularization for retain set preservation—achieves a high task aggregate score of 0.973. While the 7B variant maintained comparable forget set removal efficacy (0.964 task score), its 46\% MMLU degradation underscores the critical need for more robust mechanistic interventions. These findings suggest three key implications for machine unlearning research: First, that MLP layers in early transformer blocks serve as preferential targets for factual knowledge modification; second, that output token cross-entropy provides a more surgical intervention than full-sequence loss calculations; and third, that layer freezing thresholds must scale non-linearly with model depth to maintain utility. Future work should investigate constrained gradient updates through KL divergence to reduce the drop in model utility after unlearning. Our results cement causal mediation as a vital tool for developing compliant, adaptable LLMs that meet evolving data privacy requirements without full retraining.
\section{Acknowledgment}
We thank the anonymous reviewers for their constructive feedback. We also thank TAMU FLAIR Lab for the valuable discussions. Portions of this research were conducted with the advanced computing resources provided by Texas A\&M High Performance Research Computing.

\bibliography{custom}
\newpage
\appendix
\section{Parameter Selection for Adaptive Regularization}
\label{app:param}
To balance the preservation of retain set knowledge with the effective removal of forget set information, we introduced an adaptive regularization weight, denoted as $\alpha$, applied to the retain set loss $(\mathcal{L}^{retain})$. This weight dynamically adjusts based on the deviation of current retain set loss from its value at the start of the unlearning process. \\

\begin{figure}[h]
    \centering
    \includegraphics[width=1\linewidth]{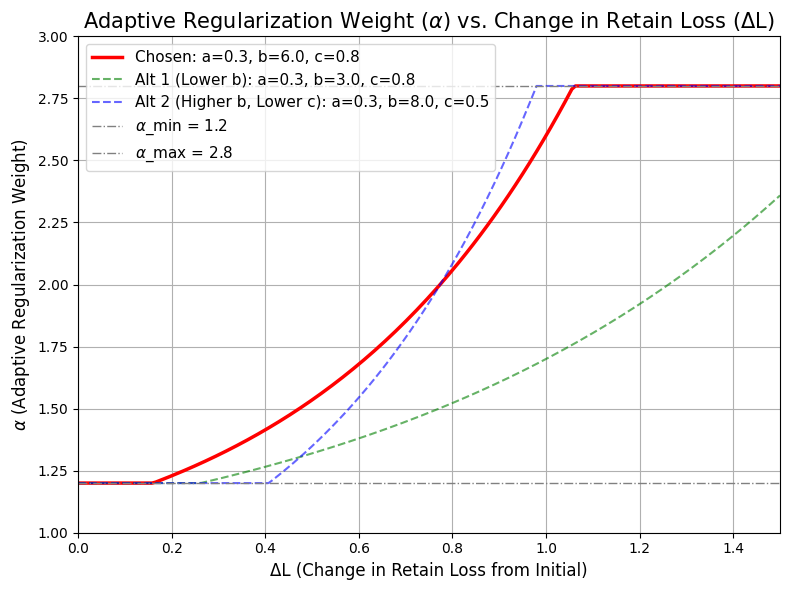}
    \caption{Visualization of the adaptive regularization function $\alpha$, plotted against the change in retain loss $\Delta L$. The chosen configuration (solid red) strongly penalizes increases in $\Delta L$ for the range of observed $\Delta L$ values, compared to blue or green.}
    \label{fig:alpha}
\end{figure}

We empirically set a=0.3, b=6, c=0.8, $\alpha_{min}=1.2$, and $\alpha_{max}=2.8$ (Figure \ref{fig:alpha}). The selected b=6 controls exponential sensitivity to $\Delta L$, strongly penalizing moderate retain loss increases to preserve performance. The offset c=0.8 and minimum clip $\alpha_{min}$ set a baseline regularization, while $\alpha_{max}$ prevents excessive strength. This configuration achieved a robust experimental balance by strongly penalizing retain set deviations while permitting effective unlearning.

\end{document}